\documentclass[journal]{IEEEtran}
\IEEEoverridecommandlockouts
\usepackage{cite}
\usepackage{amsmath,amssymb,amsfonts}
\usepackage{algpseudocode}
\usepackage{algorithm}
\usepackage{graphicx}
\usepackage{textcomp}
\usepackage{xcolor}
\usepackage{mathtools}
\usepackage{fancyhdr}
 \usepackage{multirow}
 \usepackage[a-1b]{pdfx}
\usepackage{etoolbox}

\def\BibTeX{{\rm B\kern-.05em{\sc i\kern-.025em b}\kern-.08em
    T\kern-.1667em\lower.7ex\hbox{E}\kern-.125emX}}
\usepackage{nopageno}
\renewcommand{\baselinestretch}{0.94}

\begin{document}

\pagestyle{empty}

\title{FedNET: Federated Learning for Proactive Traffic Management and Network Capacity Planning}

\author{Saroj~Kumar~Panda,~Basabdatta~Palit,~and~Sadananda~Behera%
\thanks{S. K. Panda and S. Behera are with the Department of Electronics and Communication Engineering, National Institute of Technology Rourkela, Odisha, India (corresponding author: beherasadananda@nitrkl.ac.in).}%
\thanks{B. Palit is with the Department of Information Technology, Indian Institute of Engineering Science and Technology Shibpur, Howrah, India.}%
}

\maketitle

\thispagestyle{empty}
\begin{abstract}

We propose \textbf{\textit{FedNET}}, a proactive and privacy-preserving framework for early identification of high-risk links in large-scale communication networks, that leverages a distributed multi-step traffic forecasting method. FedNET employs Federated Learning (FL) to model the temporal evolution of node-level traffic in a distributed manner, enabling accurate multi-step-ahead predictions (e.g., several hours to days) without exposing sensitive network data. Using these node-level forecasts and known routing information, FedNET estimates the future link-level utilization by aggregating traffic contributions across all source-destination pairs. The links are then ranked according to the predicted load intensity and temporal variability, providing an early warning signal for potential high-risk links. We compare the federated traffic prediction of FedNET against a centralized multi-step learning baseline and then systematically analyze the impact of history and prediction window sizes on forecast accuracy using the $R^2$ score.
Results indicate that FL achieves accuracy close to centralized training, with shorter prediction horizons consistently yielding the highest accuracy ($R^2
>0.92$), while longer horizons providing meaningful forecasts ($R^2 \approx 0.45\text{--}0.55$). 
We further validate the efficacy of the FedNET framework in predicting network utilization on a realistic network topology and demonstrate that it consistently identifies high-risk links well in advance (\textit{i.e.,} three days ahead) of the critical stress states emerging, making it a practical tool for anticipatory traffic engineering and capacity planning.
\end{abstract}


\section{Introduction}

The promise of high data rates and continuous connectivity, offered by  next-generation communication technologies, such as 5G and 6G, has triggered an increase in the demand for data-intensive applications such as augmented and virtual reality, connected healthcare, smart transportation, and large-scale IoT ecosystems~\cite{giordani2020toward,de2021survey}. These services demand high reliability and ubiquitous Quality of Service (QoS). Meeting these requirements poses a significant challenge for modern communication networks -  such as optical backbones, wireless infrastructures, and enterprise domains -  particularly under increasingly dynamic and heterogeneous traffic conditions. 

A major bottleneck in maintaining consistent QoS lies in effectively anticipating traffic growth and identifying potential high-use routes before they reach critical load thresholds~\cite{deebak2020dynamic}. In large-scale optical infrastructures, where capacity provisioning and wavelength deployment occur over longer planning cycles, predictive insights into traffic evolution are crucial for proactive capacity augmentation and resource optimization~\cite{jose2015high}. Such forecasting also helps mitigate key optical-layer challenges, including spectrum fragmentation, inefficient wavelength utilization, and connection blocking, by enabling timely defragmentation and dynamic reconfiguration, thereby ensuring seamless service continuity and efficient use of optical resources.

Traditionally, network  management frameworks have been reactive, implemented through mechanisms such as simple network management protocol (SNMP)-based polling \cite{case1990rfc1157}, threshold alarms, and software-defined networking (SDN) feedback loops \cite{feamster2014road}. These approaches are well-suited to handle failures or overloads after a network failure has occurred. However, as discussed earlier, current network scenarios are being designed to handle bandwidth-intensive real-time traffic with stringent QoS demands, which may significantly benefit from proactive and predictive traffic management. To meet these challenges, machine learning (ML)-powered traffic management has emerged as a key enabler of dynamic resource allocation, improving spectral utilization, reducing power consumption, and minimizing infrastructure overhead ~\cite{panayiotou2023survey}.

The networking architecture in recent times has also migrated from unified proprietary silos to an open, disaggregated, software-driven infrastructure, which has made distributed network management imperative. In such a scenario, the efficacy of centralized ML models, trained with data collected in a controlled setup, becomes limited. Instead, a distributed approach, such as federated learning (FL), offers a compelling alternative by enabling distributed model training at the network edge, with only aggregated model parameters shared with a central server~\cite{mcmahan2017communication, li2020federated}. It also preserves data privacy and reduces bandwidth consumption while supporting collaborative intelligence across~\cite{chen2024privacy}.

In this work, we propose \textbf{\textit{FedNET}}- a proactive, privacy-preserving framework for multi-step traffic forecasting and link utilization estimation in a distributed optical network architecture. 
We have used a \textbf{\textit{federated learning}} setup with a \textbf{\textit{Long Short Term Memory (LSTM)}}  model at its core and the \textbf{\textit{Federated Averaging (FedAvg) }} aggregation strategy \cite{li2020federated}. FL accounts for the disaggregated nature of the network, while LSTM caters to the time-series nature of the network traffic at each node. These node-level forecasts are subsequently mapped into link-level traffic estimates from the underlying routing information. By aggregating the predicted traffic contributions from all flows traversing each link, a comprehensive view of the link-level traffic is obtained. Here, \textit{link utilization} denotes the predicted traffic load carried by each link, reflecting its expected usage intensity over the forecast horizon. We then rank the links according to their predicted load intensity and variability over a future time, providing a data-driven basis for proactive capacity planning and early risk identification. 
 Link-level ranking supports physical-layer optimization tasks such as proactive load balancing, spectrum defragmentation, and impairment-aware routing by highlighting links likely to experience increased utilization. These insights will enable operators to optimize wavelength assignment, schedule reconfiguration events, and plan capacity upgrades well in advance, ensuring sustained network performance and efficient utilization of optical resources.

We benchmark the performance of FedNet with the traditional centralized multi-step training approach and establish that FL achieves prediction accuracy close to that of centralized learning (CL), while preserving data privacy and reducing communication overhead. Furthermore, we have also analyzed the influence of forecasting parameters, specifically the history window size and prediction horizon, on the accuracy of traffic prediction, using the $R^2$ score as a benchmark. Our results demonstrate how these parameters affect the reliability of utilization forecasting and the robustness of link-level ranking. 

In summary, the key contributions of this paper are as follows.
\begin{itemize}
\item To the best of the authors' knowledge, \textbf{FedNET} is the first federated learning framework for multi-step traffic prediction in optical networks. Our approach enables distributed forecasting across disaggregated domains without requiring the exchange of raw traffic data.
\item We have established through a benchmarking study that FL achieves comparable accuracy (measured by the $R^2$ score), similar to a traditional centralized multi-step training baseline, while offering advantages in privacy preservation and reduced communication overhead.
\item We have also proposed a novel heuristic that translates node-level multi-step forecasts into 
link-level utilization scores by leveraging the routing knowledge, enabling proactive identification of at risk links well before any critical states emerge.
\item We systematically analyze the effects of history and prediction window sizes on both forecast accuracy and the robustness of link-level utilization score through extensive simulations on a realistic optical network topology.
\end{itemize}

Through extensive simulations, we have observed that the proposed \textit{FedNET} framework achieves accurate multi-step traffic prediction, with $R^{2}$ scores exceeding $0.92$ for short prediction horizons and remaining in the range of $0.45$–$0.55$ for up to three days into the future. At the same time, the framework is able to consistently identify high-risk links well before they approach critical states, thereby providing a reliable early-warning mechanism for proactive traffic engineering.

The remainder of this paper is structured as follows. Section \ref{related} reviews the related work, while Section \ref{fednet} provides an overview of the proposed approach along with the design and implementation details. Training and evaluation are presented in Section \ref{training}. Section \ref{high-risk-link} introduces the proposed method for identifying high-risk links, and concluding remarks are provided in Section \ref{conclusions}.

\section{Related Works}
\label{related}
\begin{figure*}[t]
\centering
\includegraphics[width = 0.75\textwidth, trim = {4cm 10cm 4cm 0cm}]{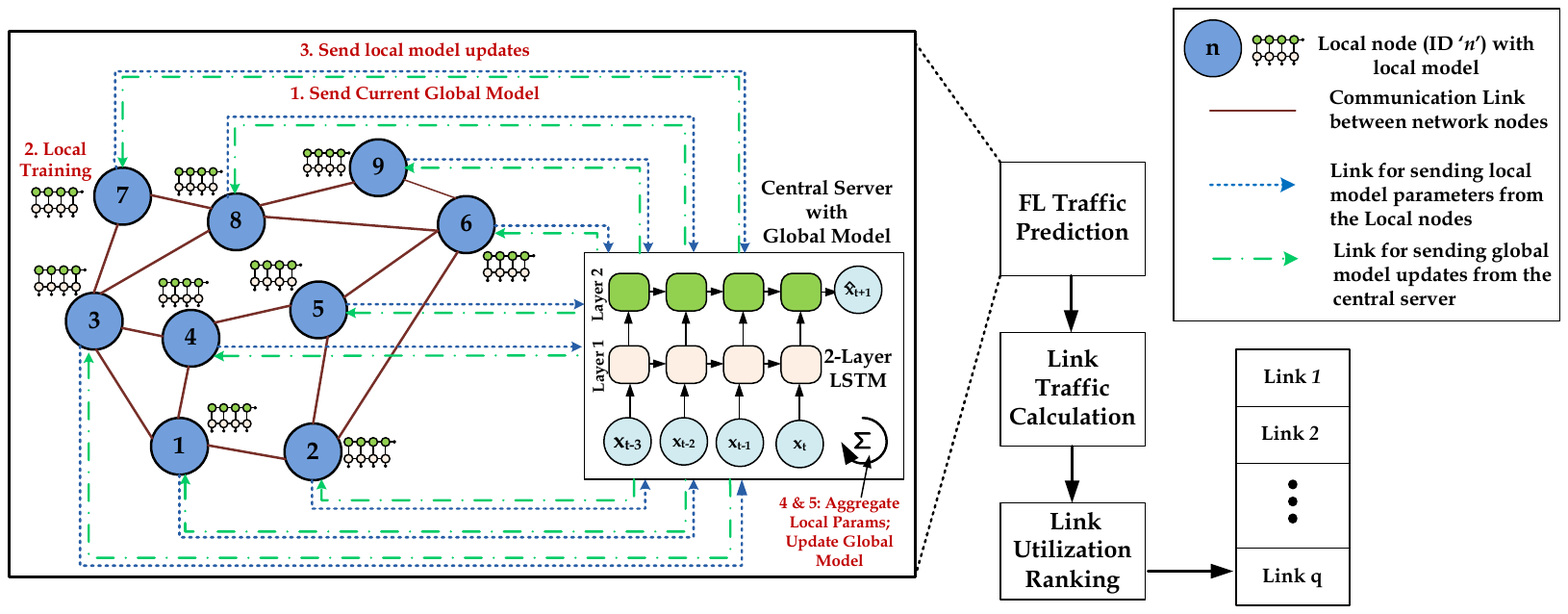}
\caption{The flowchart of the FedNET architecture illustrates its implementation over the 9-node BRAIN network topology.}
\label{fig:network_topology}
\end{figure*}

Network traffic prediction is fundamentally a time-series forecasting problem, where historical observations are used to anticipate future demands~\cite{panayiotou2023survey}. Early studies demonstrated the predictability of traffic patterns using centralized statistical models such as Auto Regressive Integrated Moving Average (ARIMA) and its extensions~\cite{alghamdi2019forecasting}. While effective over short horizons, these methods assume linearity and often fail to generalize for long-term dependencies and the nonlinear characteristics of high-speed traffic. With the rise of deep learning, models such as Recurrent Neural Networks (RNNs),  LSTM networks, and Gated Recurrent Units (GRUs) have been widely applied~\cite{tedjopurnomo2020survey}, offering improved accuracy by learning temporal correlations in traffic sequences. However, both statistical and deep learning approaches typically rely on centralized data collection at a controller, raising concerns of privacy, scalability, and communication overhead.

To mitigate these limitations, FL has emerged as a distributed paradigm that trains models locally at clients while sharing only aggregated updates with a central coordinator. Extensive research has validated FL as a privacy-preserving alternative to centralized training, addressing challenges such as communication efficiency, model scalability, and concept drift~\cite{mcmahan2017communication,li2020federated,mcmahan2016federated,manias2021concept,konevcny2016federated,li2020bandwidth,li2021scalable,ruan2020low}. For instance, \cite{drainakis2023centralized,9306745, 9605846,9748424,Chen:19} report that FL-based models for traffic prediction closely match centralized baselines, while reducing communication costs and avoiding raw data exchange across domains.
Recent works have also explored fairness-aware FL~\cite{pandaskffl,panda2025federated}, showing that collaborative training can balance prediction accuracy across heterogeneous clients and improve service-level fairness. While these studies demonstrate the feasibility of FL for distributed traffic prediction, most focus on short-term forecasts and emphasize performance parity with centralized methods rather than novel use cases.

Despite growing interest, the application of FL to optical networks remains underexplored. Optical infrastructures pose unique challenges due to their multi-layer architectures, disaggregated vendor ecosystems, and extremely high data rates. In~\cite{behera2024federated}, a federated framework was shown to achieve accuracy comparable to centralized models while improving privacy, reducing latency, and lowering bandwidth usage through distributed training. In~\cite{ciceri2022federated}, FL was applied to next-generation passive optical networks, demonstrating feasibility in access-layer settings. Beyond optical transport, FL has been investigated for intelligent traffic prediction~\cite{liu2023multilevel}, lightweight IoT frameworks~\cite{chen2024privacy}, and 6G-integrated optical systems~\cite{anastasopoulos2025demonstration}, while ML techniques have been applied to address traffic dynamicity in optical access networks~\cite{wong2023machine}. Other works highlight challenges in failure management~\cite{musumeci2025failure} and explore privacy-preserving schemes such as vertical FL for failure localization across multi-vendor environments~\cite{ibrahimi2024vertical}. Dynamic link load prediction has also been explored for failure recovery~\cite{knapinska2023link}, though not in a federated setting.

Although FL has shown promise for traffic prediction across diverse networking contexts, most existing studies remain limited to short-term forecasts and stop at node-level insights. Long-horizon forecasting is challenging due to error accumulation and bias amplification in federated settings~\cite{9528995,10.1145/3627345.3627356}. As a result, the critical task of translating node-level predictions into actionable link-level utilization metrics remains largely underexplored. To bridge this gap, we propose FedNET, a novel FL-driven framework that performs multi-step node-level forecasting (up to three days ahead), maps these predictions to link-level utilization via routing paths, and ranks links according to their predicted load and variability. This provides operators with an early warning system for proactive capacity planning and operational decision-making.

\section{Proposed Approach and Methodology}
\label{fednet}
The FedNET architecture uses FL to predict node-level traffic in a distributed network environment. It subsequently maps this node-level traffic into link-level traffic predictions to rank links based on their predicted load and variability.

In this section, we explain how we have used FL for traffic prediction in a realistic network, to design FedNET. We first provide an overview of the approach in Section \ref{sec:approach_overview}, highlighting how FedNET predicts multi-step traffic patterns that can subsequently be utilized for high-risk link identification. Following this in Section \ref{sec:federated_traffic_prediction}, we describe the methodology adopted to implement FedNET, including the dataset preparation, client-side training, and server-side aggregation. The algorithm for the high-risk link identification is explained in Section \ref{high-risk-link}.

\subsection{Approach Overview of FedNET}\label{sec:approach_overview}

As outlined in Fig.\ref{fig:network_topology}, FedNET first predicts the node level traffic using FL. A central server (typically an SDN orchestrator) maintains a global LSTM model, which is iteratively trained through periodic updates from distributed cohorts. The process begins with a one-time initialization of the global model at the central server and a subsequent distribution of the parameters (Step 1). Upon receiving the global LSTM model, each local client trains it using their local traffic data (Step 2). The updated weights are then transmitted back to the server (Step 3), where they are aggregated and the global model is updated (Steps 4–5). The global model sends the updated weights back to the clients at the beginning of the next round. This cycle continues until (i) it reaches convergence or, (ii) it meets a predefined stopping criterion.

The trained global model, once aggregated, is redistributed to the clients, where it is used to generate multi-step node-level forecasts. Using routing information, these predictions are mapped to links, and a link-utilization score is derived to highlight links with higher predicted load and variability.

\subsection{The Federated Multi-Step Traffic Prediction}\label{sec:federated_traffic_prediction}
 Traffic in a real-world network can flow from any node to any other node. Our objective is to predict the temporal variation of this traffic at each node over a prediction window $p$ using the traffic of the past $h$ time instants. As the network traffic constitutes time-series samples, we predict the traffic at each node using an LSTM model combined with the \textit{FedAvg} aggregation strategy~\cite{li2020federated}. 
 
  \subsubsection{Details of the Dataset and Pre-processing}
 \label{dataset}
 To establish the efficacy of FL in predicting optical network traffic,  we have used the BRAIN network topology, which is a well-known 9-node, 14-link reference network from the SNDlib repository~\footnote{https://sndlib.put.poznan.pl/brain.overview.action, accessed on \today}
as depicted in 
Figure~\ref{fig:network_topology}.  The dataset $D_k$,  corresponding to node $k$ in the SNDlib, contains the aggregated bit-rate in $1$-hour intervals. The hourly data undergoes a multi-stage pre-processing pipeline in our work as follows. 
\begin{enumerate}[]
    \item \textit{\textbf{Reducing the time scale:}} We first average the original hourly traffic measurements from SNDlib over non-overlapping 6-hour windows, thereby averaging over short-term fluctuations while highlighting longer-term trends in the traffic patterns.
    \item \textit{\textbf{Outlier Removal: }}Subsequently, to address the anomalies in the data, we apply an interquartile range (IQR)-based outlier detection, i.e., any data point lying outside the range defined by $Q_1-1.5\times \text{IQR}$ and $Q_3+1.5\times \text{IQR}$ (where, $Q_1$ and $Q_3$ are the $20^{th}$ and $80^{th}$ percentiles, respectively) is marked as an outlier. These outliers are then replaced using mean imputation, ensuring continuity in the time series.
    \item \textbf{\textit{Moving-average Filtering: }}Next, a moving-average smoothing with a window size of $28$ samples is applied to suppress residual noise and further stabilize the traffic profiles for FL.
    \item \textbf{\textit{Normalization:}}
    Each dataset $D_k$ is further normalized using the StandardScaler.
\end{enumerate}
  
After  preprocessing, all the $9$ nodes yield processed datasets
\mbox{$\{\tilde{D}_k\}_{k=1}^9 = [845, 945, 1445, 1047, 1445, 645, 547, 667, 967]$} samples.

For each client $k$, the input to the model comprises of the two vectors $\mathbf{x}^k$ and $\mathbf{y}^k$, where $\mathbf{x}^k$ is a sequence of network traffic data that spans the previous $h$ time steps and $\mathbf{y}^k$ represents the ground truth of the future traffic over $p$ time steps, s.t.,

\begin{equation*}
    \mathbf{x}^{k} = [x_{t-h+1}^{k}, \dots, x_{t}^{k}], \quad \mathbf{y}^{k} = [x_{t+1}^{k}, \dots, x_{t+p}^{k}].
\end{equation*}

\noindent Here, $x_{i}^k$ denotes the traffic measurement at node
$k$ at the time step $i$. The history window and the prediction horizons take values such as $h,p\in\{1,4,8,12\}$. These values of $h,p$ are selected to reflect different forecasting ranges under the previously mentioned $6$-hour sampling interval, i.e., $p=12$ corresponds to a prediction window $3$ days ahead, while smaller values capture short- and medium-term dynamics. This setup allows us to examine how accuracy changes across timescales while keeping experiments computationally feasible.

\subsubsection{Client-Side Local Training using LSTM for Traffic Prediction as Time-series data}
Each local client $k$ receives the global LSTM model from the central server in each communication round and trains it using their local data  $\{\mathbf{x}^k,\mathbf{y}^k\}\in D_k$. The LSTM model is trained to learn the future throughput $\mathbf{\hat{y}}^{k} = [\hat{x}_{t+1}^k, \hat{x}_{t+2}^k,..,\hat{x}_{t+p}^k]$, over the time instants ${t+1, t+2, \cdots t+p}$, 
by minimizing the loss function $\mathcal{L}_k(w_k)$, such that, 
\begin{equation}
    \underset{w_k}{min} \ \mathcal{L}_k(w_k) = \frac{1}{p}\sum_{i=1}^{p}\left({x}^k_{t+i}-\hat{{x}}^k_{t+i}\right)^2
\end{equation}

Comprehensive discussions of FL frameworks and LSTM architectures can be found in \cite{mcmahan2017communication} \cite{Kim_Jernite_Sontag_Rush_2016}. 

 \subsubsection{Server-side Aggregation}
 The global objective in FL is to minimize the global loss function $\mathcal{L}(w)$ i.e.,
 \begin{equation}
     \min_{w} \mathcal{L}(w) = \sum_{k=1}^{K} \frac{|D_k|}{N} \mathcal{L}_k(w)
\end{equation}
where:
\begin{itemize}
    \item $\mathcal{L}_k(w)$: Local loss for client $k$
    \item $|D_k|$: Number of samples for client $k$
    \item $N$: Total samples across all clients ($N = \sum_k |D_k|$)
    \item $w$: Global model parameters
\end{itemize}
At the server, the global model is updated by aggregating the weights received from all clients after local training, using the \textit{FedAvg} aggregation strategy. Let $w_k^r$ denote the model weight from client $k$ at communication round $r$. With $K$ clients, the server computes a weighted average of client models:
\[
 w^{r+1} = \sum_{k=1}^{K} \alpha_k .w_k^r,
 \hspace{0.3cm}
 \alpha_k = \frac{|D_k|}{N}
\]
where $\alpha_k$ reflects the relative size of each client’s dataset. The updated global model $w^{r+1}$ is then redistributed to all clients for the next round of local training, ensuring iterative improvement and consistency across the federation.

\section{Evaluation}
\label{training}
In this section, we first explain the implementation of the FL based traffic prediction in the optical networks in Section \ref{sec:Implementation} and then the results in Section \ref{sec:results}.

\subsection{Implementation} \label{sec:Implementation}
In this section, we outline the implementation details including the hyperparameter configurations for federated learning, and the computational resources used for evaluation.
\subsubsection{Hyperparameters for FL}

For local training at each client, the  dataset $D_k$, is split into $70\%$ training and $30\%$ testing sets, with time-series samples generated through sliding window processing. Furthermore, the final $20\%$ of the training dataset is reserved for validation. Local training employed a batch size of $256$, a learning rate of $0.001$, and ran for one epoch. The \textit{ FedAvg} aggregation at the server, for each combination of $h$ and $p$, is performed over $50$ communication rounds. The LSTM architecture comprises of two hidden layers with $0.2$ dropout and a RepeatVector layer, compiled with the Adam optimizer to minimize the mean squared error (MSE). The model was defined on the server and distributed to all clients to ensure architectural consistency.

To evaluate forecasting performance, we have tested the FL module for all combinations of $h,p\in\{1,4,8,12\}$ intervals. This evaluates how the historical depth affects the prediction accuracy across immediate ($p = 1$) to extended prediction horizons ($p = 12$), revealing the model's capability to capture multi-scale temporal patterns in network traffic.
\subsubsection{Compute resources}
The experimental framework leverages Python 3.9.0 and TensorFlow 2.10.1, executed on hardware featuring an Intel® Core™ i5-10300H processor (2.50 GHz), 16 GB RAM, and an NVIDIA GeForce GTX 1650 GPU. 
\subsection{Results}\label{sec:results}
In this section, we first establish the efficacy of FL in predicting the network traffic by benchmarking its performance against the traditional centralized learning method. We then use a parametric analysis to show the variation of the $R^2$ score and MSE loss with respect to the history window and the prediction horizon.
\subsubsection{Comparative Analysis of FL with Centralized Learning}
\label{comparison}
\begin{table*}[ht]
\centering
\caption{$R^{2}$ scores of individual clients for centralized vs. federated learning under two window settings: $h,p=1$ and $h,p=12$.}
\label{table:comparision}

\begin{tabular}{|c|c|c|c|c|c|c|c|c|c|c|c|c|}
\hline
\textbf{h} & \textbf{p} & \textbf{Model} & \textbf{Client 1} & \textbf{Client 2} & \textbf{Client 3} & \textbf{Client 4} & \textbf{Client 5} & \textbf{Client 6} & \textbf{Client 7} & \textbf{Client 8} & \textbf{Client 9} & \textbf{Avg} \\ \hline

\multirow{2}{*}{\textbf{1}} & \multirow{2}{*}{\textbf{1}} & \textbf{CL} & $0.973$ & $0.993$ & $0.983$ & $0.987$ & $0.960$ & $0.983$ & $0.978$ & $0.984$ & $0.967$ & $0.979$ \\ \cline{3-13} 
&& \textbf{FL} & $0.973$ & $0.993$ & $0.983$ & $0.982$ & $0.958$ & $0.984$ & $0.977$ & $0.983$ & $0.968$ & $0.978$ \\ \hline

\multirow{2}{*}{\textbf{12}} & \multirow{2}{*}{\textbf{12}} & \textbf{CL} & $0.651$ & $0.770$ & $0.609$ & $0.796$ & $0.569$ & $0.712$ & $0.841$ & $0.781$ & $0.766$ & $0.722$\\ \cline{3-13} 
&& \textbf{FL} & $0.635$ & $0.737$ & $0.632$ & $0.740$ & $0.522$ & $0.676$ & $0.852$ & $0.740$ & $0.746$ & $0.698$ \\ \hline

\end{tabular}

\end{table*}

In this work, we have treated the centralized learning (CL) method as the baseline. The individual test sets from all clients are aggregated into a unified test dataset. Similarly, remaining $70\%$ is split into $80\%$ for centralized training and $20\%$ for validation. This ensures a fair and consistent comparison between the two approaches.
Table~\ref{table:comparision} presents the $R^{2}$ scores for the individual clients under both FL and CL, considering two representative extremes of the history window ($h$) and prediction window ($p$): $(h,p)=(1,1)$ and $(h,p)=(12,12)$. The results reveal only marginal differences in prediction accuracy, with FL closely matching CL across all clients. As anticipated, CL achieves slightly higher scores due to its access to the complete dataset during training. Importantly, FL maintains robustness across heterogeneous clients, demonstrating its ability to deliver consistent accuracy despite data remaining non-identically distributed across the client cohorts. This establishes FL as a competitive and privacy-preserving alternative to CL in large-scale, distributed network environments.
\subsubsection{Performance Evaluation of FL}
\begin{figure}[h]
\centering
\includegraphics[width=0.45\textwidth,trim={0.5cm 0.5cm 1cm 0.5cm}]{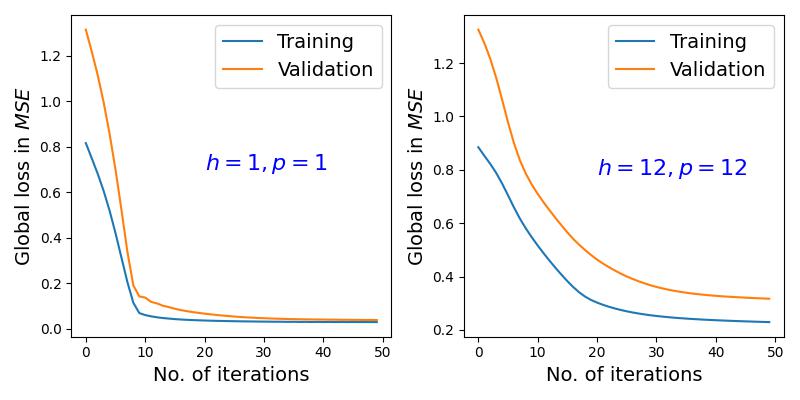} 
\caption{Global loss across all clients under two window settings: $h,p=1$ and $h,p=12$.}
\label{learning_curves}
\end{figure}

Figure~\ref{learning_curves} illustrates the performance of the FL-based traffic prediction in terms of the MSE loss, by varying the history window $h$ and the prediction horizon $p$ from the minimal ($h,p=1$) to the maximal ($h,p=12$) window configurations, highlighting the critical role of these window sizes on the prediction accuracy.  With $h,p=1$, it is observed that both training and
validation converge to an MSE loss that is close to zero, indicating highly accurate short-term prediction. In contrast, for $h,p=12$, both losses also converge; however, they exhibit a slight deviation between training and validation, primarily due to the increased uncertainty associated with longer prediction horizons. Other configurations showed similar behavior, varying between the minimum and the maximum window sizes as discussed next.

\begin{figure}[!h]
\centering
\includegraphics[width=0.5\textwidth, trim={0cm 0.5cm 0cm 1.5cm}]{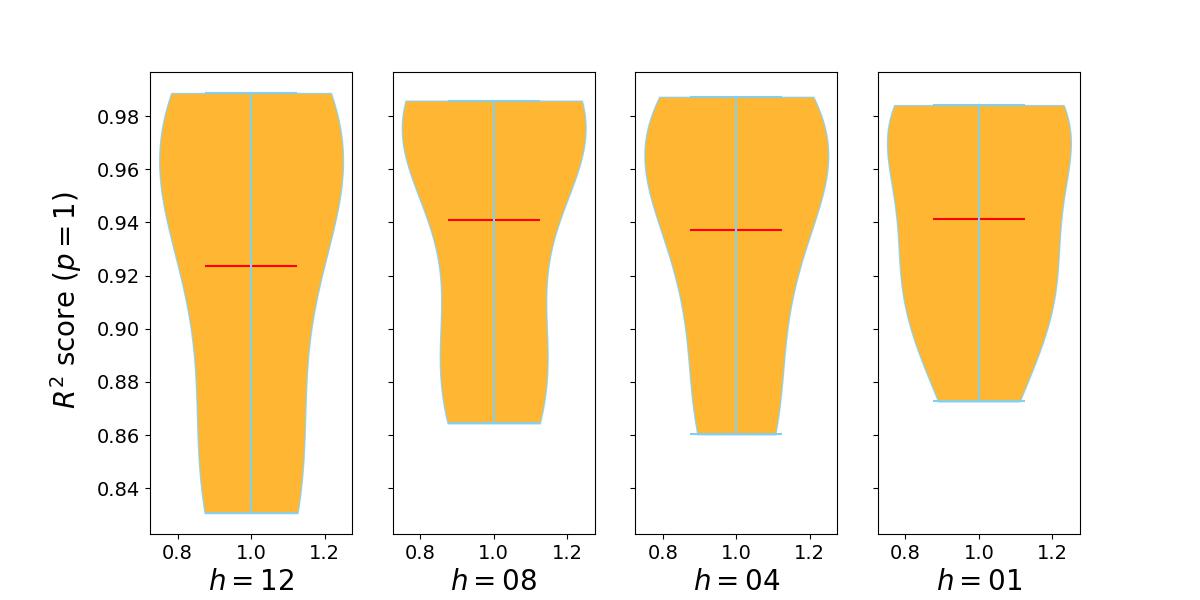}
\caption{Distribution of Client $R^2$ scores for varying history window sizes.}
\label{varying_history_size}
\end{figure}

\begin{figure}[!h]
\includegraphics[width=0.5\textwidth, trim={0cm 0.5cm 0cm 1.5cm}]{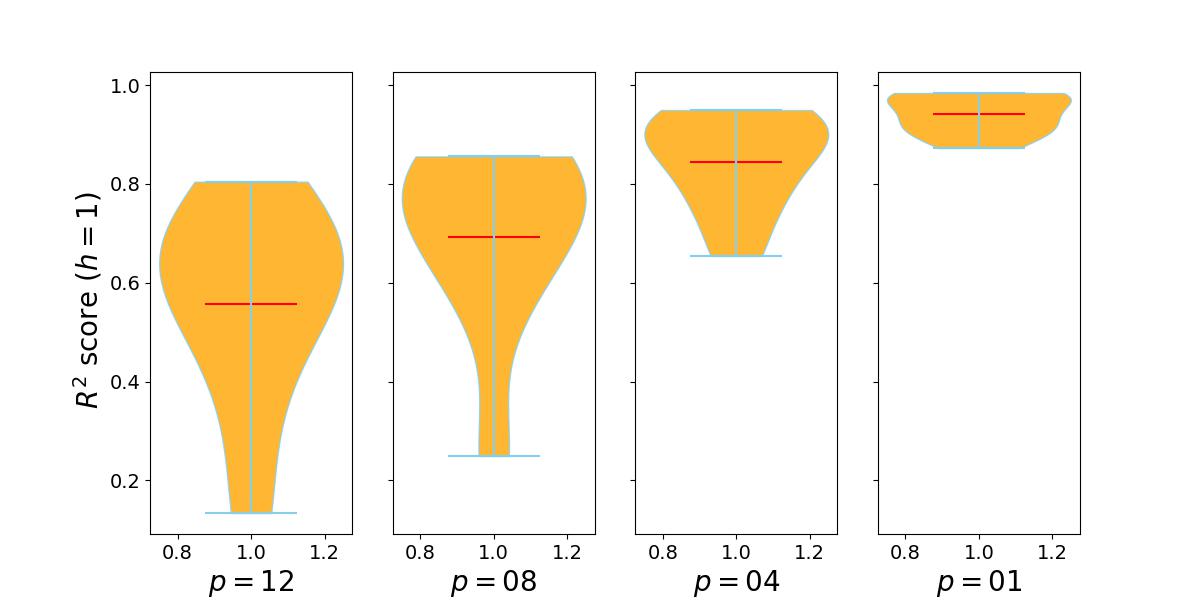}
\caption{Distribution of Client $R^2$ scores for varying prediction horizons with a history window size of $h=1$.}
\label{varying_prediction_size}
\end{figure}

Figures~\ref{varying_history_size} and ~\ref{varying_prediction_size} illustrate the effect of the temporal window sizes on the model performance, using violin plots that demonstrate the distribution of clients' $R^{2}$ scores. In Fig.~\ref{varying_history_size}, the history window length varies ($h = 1, 4, 8, 12$) while keeping the prediction window constant ($p = 1$); conversely, Fig.~\ref{varying_prediction_size} explores different prediction windows $(p = 1, 4, 8, 12)$ with a fixed history length ($h = 1$). The red horizontal lines in each violin plot correspond to the mean $R^{2}$ scores shown numerically in Table~\ref{table:avg_r2_scores}, which provide comprehensive average scores for all $(h,p)$ pair combinations. It is observed from Fig.~\ref{varying_history_size} that the highest mean $R^2$ score and the lowest variance are observed when $h=1,p=1$. The variance increases while the mean decreases with an increase in the history window. Similarly, the mean decreases while the variance increases with an increase in the prediction window. Thus, two major insights emerge: (i) an overall inverse relationship between window size and model accuracy, where increasing either history or prediction length generally lowers $R^{2}$ scores, and (ii) a steeper decline in performance along the prediction axis $(p = 1$ → $12)$ than along the history axis $(h = 1$ → $12)$. This asymmetry indicates that the model is more sensitive to the length of the forecasting horizon than to the amount of historical context. Such behavior reflects the autoregressive nature of the system, where error propagation in multi-step predictions contributes more to degradation than does the challenge of extracting patterns from longer input sequences. These findings emphasize the importance of carefully tuning the prediction window size to maintain forecast reliability, while suggesting that the model remains relatively robust to longer history inputs.
\footnotesize
\begin{table}[h]
\centering
\caption{Average $R^{2}$ scores for different combinations of $(h,p)$.}
\label{table:avg_r2_scores}
\begin{tabular}{|c|c|c|c|c|c|}
\hline
\multicolumn{2}{|c|}{} & \multicolumn{4}{c|}{\textbf{Prediction Window Size ($p$)}} \\ 
\cline{3-6} 
\multicolumn{2}{|c|}{\multirow{-2}{*}{\textbf{$R^2$ Score}}} & 
\textbf{12} & \textbf{8} & \textbf{4} & \textbf{1} \\ 
\hline
\multicolumn{1}{|c|}{\multirow{4}{*}{\textbf{History Window Size ($h$)}}} & \textbf{12} & 0.45 & 0.62 & 0.80 & 0.92 \\ 
\cline{2-6} 
\multicolumn{1}{|c|}{} & \textbf{8} & 0.50 & 0.67 & 0.83 & 0.94\\ 
\cline{2-6} 
\multicolumn{1}{|c|}{} & \textbf{4} & 0.50 & 0.70  & 0.84 & 0.94 \\ 
\cline{2-6} 
\multicolumn{1}{|c|}{} & \textbf{1} & 0.58 & 0.70 & 0.85 &  0.941 \\ 
\hline
\end{tabular}
\end{table}
\normalsize

\section{High-risk Link Identification in FedNET}
\label{high-risk-link}
The FL-based traffic prediction model, discussed so far, constitutes the heart of the FedNET framework. The node-level traffic forecasts, obtained from the FL prediction engine, are subsequently used to estimate link-level utilization and identify links with higher predicted load and variability by combining them with routing information to estimate link utilization and rank the links based on their predicted risk. In this section, we first describe the mapping of predicted node-level traffic to link-level flows, followed by the computation of link scoring based on predicted utilization. We then analyze how forecasting parameters such as history and prediction windows affect these scores.

\subsection{Node-Level to Link-Level Traffic Mapping}

Let the network be modeled as a graph $G = (\mathcal{K}, \mathcal{L})$ where:
\begin{itemize}
    \item $\mathcal{K}$ : Set of all nodes, corresponding to the clients, in the network, indexed by $k$. $|\mathcal{K}|$ = total number of nodes in the network.
    \item $\mathcal{L}$ : Set of links, connecting the nodes, indexed by $\ell$.
    \item${T}$: Length of the time-series at the nodes.  Since the datasets used at different nodes have varying lengths, the resulting prediction sequences are not uniform across nodes. To ensure temporal alignment, all prediction sequences are truncated to match the shortest available length, ${T}$. This ensures that traffic mapped onto links corresponds to the same timestamps throughout the network, allowing for consistent evaluation of link-level utilization. The time steps are indexed by $t$.
    
    \item $\mathcal{U}$: Set of time sequences, which are obtained by sliding the history window $h$ and prediction horizon $p$ over the time-series data, given by $|\mathcal{U}| =   {T}- (h+p)+1$. Time sequences are indexed by $m$.
    \item  $\hat{{x}}^k_m(t)$ : Predicted traffic at node  $k \in K $ at time step $t$ for $m^{th}$ sequence.
    
    \item $\mathcal{P}^{(s,d)}$ : Shortest path (minimum hop count) between source node $s$ and destination node $d$, represented as a subset of links from $\mathcal{L}$.
\end{itemize}
\begin{algorithm}[t]
\caption{Identification of High-Load Links Using Predicted Traffic}
\label{alg:congested_link_identification}
\begin{algorithmic}[1]
\item[]  \textbf{Input:} Predicted node traffic $\hat{x}^k_{m}(t)$  for all nodes $ k \in \mathcal{K} $, sequences $ m \in \mathcal{U} $, and time steps $ t \in \mathcal{T}=\{1, \cdots, T\} $.
\item[]  \textbf{Output:} Ranked list of top $ q $ highly utilized links.

\ForAll{link $ \ell \in \mathcal{L} $}
    \ForAll{sequence $ m = 1 $ to $ |\mathcal{U}| $}
        \ForAll{time step $ t = 1 $ to $ T $}.
            \ForAll{$ (s, d) \in |\mathcal{K}| \times |\mathcal{K}|, s \ne d $}
                \State Compute per-pair traffic: $ \hat{x}^{(s,d)}_{m}(t)$ from (\ref{sd_traf}).
                \State Find shortest path $ \mathcal{P}^{(s,d)} $ from $ s $ to $ d $.
                \ForAll{link $ \ell \in \mathcal{P}^{(s,d)} $}
                    \State Accumulate link traffic: 
                    $ \tau_{\ell, m}(t) $ from (\ref{link_traf}).
                \EndFor
            \EndFor  
        \EndFor
        \State Compute $ \mu_{\ell, m}$ and $ \sigma_{\ell, m}$ for link $l$ and sequence $m$.
        
    \EndFor
    \State Aggregate across sequences:  $ \bar{\mu}_\ell$ from (\ref{eq:mu_l}) and $ \bar{\sigma}_\ell$ from (\ref{eq:sigma_l}).
    \State Compute link utilization score: $\zeta_\ell$ from (\ref{eq:utilization_scores}).
\EndFor

\State Rank all links in descending order of $ \zeta_\ell $
\State \Return Top $ q $ links with highest link utilization scores
\end{algorithmic}
\end{algorithm}
The algorithm for node to link level mapping and for detecting the highly utilized links across network  is outlined in Algorithm~\ref{alg:congested_link_identification}. 
  We consider all ordered source-destination pairs $(s,d) \in \mathcal{K} \times \mathcal{K}$, where $s \ne d$ . For each source node $s$, it is assumed that its predicted traffic $\hat{x}^s(t)$ is evenly distributed across all $|\mathcal{K}|-1$ possible destination nodes (refer line 5 of Algorithm \ref{alg:congested_link_identification}). Thus, the traffic from node  $s$  to each node $d \ne s$ for sequence $m$ at time stamp $t$ is given by:

\begin{equation}
\label{sd_traf}
\hat{x}^{(s,d)}_{m}(t) = \frac{\hat{x}^s_m(t)}{|K| - 1}, \hfill \forall d\in\mathcal{K}\backslash s
\end{equation}

\begin{table*}[ht]
\centering
\caption{Top-6 Highly Utilized Links of BRAIN network topology of Fig. \ref{fig:network_topology} and the corresponding predicted and actual link utilization scores for different combinations of $(h,p)$.}
\label{table:top5_links_grid}
\renewcommand{\arraystretch}{1.2}
\begin{tabular}{|c|c|c|c|c|c|}
\hline

\multirow{2}{*}{\textbf{$h$}} & \multirow{2}{*}{\textbf{$p$}} & \multicolumn{2}{c|}{\textbf{Top-6
Highly Utilized Links ($L_{l}$)}} & \multicolumn{2}{c|}{\textbf{Link Utilization Scores ($\times10^6$)}} \\
\cline{3-6}
 & & \textbf{Actual} & \textbf{Predicted} & \textbf{Actual} & \textbf{Predicted} \\
\hline

\multirow{4}{*}{12} 
& 12 
& $L_{38}, L_{43}, L_{45}, L_{68}, L_{34}, L_{31}$
& $L_{38}, L_{43}, L_{45}, L_{34}, L_{31}, L_{26}$ & 1.00, 0.72, 0.72, 0.55, 0.52, 0.52
& 1.00, 0.79, 0.77, 0.54, 0.54, 0.51\\
& 8  
& $L_{38}, L_{45}, L_{43}, L_{68}, L_{34}, L_{31}$
& $L_{38}, L_{43}, L_{45}, L_{34}, L_{31}, L_{68}$
& 1.00, 0.71, 0.70, 0.54, 0.53, 0.53 
& 1.00, 0.77, 0.76, 0.53, 0.53, 0.49\\
& 4  
& $L_{38}, L_{45}, L_{43}, L_{68}, L_{34}, L_{31}$
& $L_{38}, L_{43}, L_{45}, L_{68}, L_{34}, L_{31}$
& 1.00, 0.69, 0.67, 0.53, 0.52, 0.52
& 1.00, 0.81, 0.80, 0.53, 0.50, 0.50\\
& 1  
& $L_{38}, L_{45}, L_{43}, L_{68}, L_{89}, L_{34}$ 
& $L_{38}, L_{45}, L_{43}, L_{68}, L_{89}, L_{34}$
& 0.50, 0.39, 0.37, 0.30, 0.23, 0.23 
& 0.50, 0.39, 0.37, 0.30, 0.23, 0.23
\\ \hline

\multirow{4}{*}{8} 
& 12 
& $L_{38}, L_{45}, L_{43}, L_{68}, L_{34}, L_{31}$ 
& $L_{38}, L_{43}, L_{45}, L_{34}, L_{31}, L_{68}$
& 1.00, 0.72, 0.72, 0.55, 0.53, 0.53 
& 1.00, 0.81, 0.78, 0.53, 0.53, 0.49\\
& 8  
& $L_{38}, L_{45}, L_{43}, L_{68}, L_{34}, L_{31}$
& $L_{38}, L_{43}, L_{45}, L_{26}, L_{68}, L_{34}$
& 1.00, 0.71, 0.70, 0.54, 0.53, 0.53 
& 0.99, 0.87, 0.81, 0.54, 0.52, 0.52\\
& 4  
& $L_{38}, L_{45}, L_{43}, L_{34}, L_{31}, L_{68}$
& $L_{38}, L_{43}, L_{45}, L_{34}, L_{31}, L_{68}$
& 1.00, 0.70, 0.68, 0.52, 0.52, 0.52 
& 1.00, 0.86, 0.81, 0.51, 0.51, 0.49\\
& 1  
& $L_{38}, L_{45}, L_{43}, L_{68}, L_{34}, L_{31}$ 
& $L_{38}, L_{45}, L_{43}, L_{68}, L_{89}, L_{34}$
& 0.50, 0.39, 0.37, 0.29, 0.23, 0.23 
& 0.50, 0.39, 0.37, 0.29, 0.23, 0.23
\\ \hline

\multirow{4}{*}{4} 
& 12 
& $L_{38}, L_{43}, L_{45}, L_{68}, L_{34}, L_{31}$ 
& $L_{38}, L_{43}, L_{45}, L_{34}, L_{31}, L_{68}$
& 1.00, 0.72, 0.72, 0.54, 0.53, 0.53
& 1.00, 0.77, 0.75, 0.53, 0.53, 0.48\\
& 8  
& $L_{38}, L_{45}, L_{43}, L_{68}, L_{34}, L_{31}$
& $L_{38}, L_{43}, L_{45}, L_{34}, L_{31}, L_{68}$
& 1.00, 0.71, 0.70, 0.53, 0.53, 0.53 
& 1.00, 0.85, 0.78, 0.53, 0.53, 0.50\\
& 4  
& $L_{38}, L_{45}, L_{43}, L_{34}, L_{31}, L_{68}$ 
& $L_{38}, L_{43}, L_{45}, L_{26}, L_{68}, L_{34}$
& 1.00, 0.69, 0.67, 0.52, 0.52, 0.51 
& 0.95, 0.87, 0.78, 0.54, 0.52, 0.52\\
& 1  
& $L_{38}, L_{45}, L_{43}, L_{68}, L_{34}, L_{31}$ 
& $L_{38}, L_{45}, L_{43}, L_{68}, L_{34}, L_{31}$
& 0.50, 0.39, 0.37, 0.29, 0.23, 0.23
& 0.50, 0.39, 0.37, 0.29, 0.23, 0.23
\\ \hline

\multirow{4}{*}{1} 
& 12 
& $L_{38}, L_{43}, L_{45}, L_{68}, L_{34}, L_{31}$ 
& $L_{38}, L_{43}, L_{45}, L_{34}, L_{31}, L_{68}$
& 1.00, 0.73, 0.72, 0.54, 0.53, 0.53 
& 1.00, 0.76, 0.75, 0.54, 0.54, 0.46\\
& 8  
& $L_{38}, L_{45}, L_{43}, L_{68}, L_{34}, L_{31}$ 
& $L_{38}, L_{43}, L_{45}, L_{34}, L_{31}, L_{68}$
& 1.00, 0.71, 0.70, 0.53, 0.53, 0.53
& 1.00, 0.81, 0.77, 0.54, 0.54, 0.51\\
& 4  
& $L_{38}, L_{45}, L_{43}, L_{34}, L_{31}, L_{68}$ 
& $L_{38}, L_{43}, L_{45}, L_{68}, L_{26}, L_{34}$
& 1.00, 0.69, 0.67, 0.52, 0.52, 0.51
& 0.94, 0.87, 0.77, 0.55, 0.55, 0.53\\
& 1  
& $L_{38}, L_{45}, L_{43}, L_{68}, L_{34}, L_{31}$ 
& $L_{38}, L_{45}, L_{43}, L_{68}, L_{34}, L_{31}$
& 0.50, 0.38, 0.37, 0.29, 0.23, 0.23 
& 0.50, 0.39, 0.37, 0.29, 0.23, 0.23
\\ \hline
\end{tabular}
\end{table*}

This traffic is routed along the shortest path $\mathcal{P}^{(s,d)}$. The predicted traffic on any link $\ell$ at time $t$ for sequence $m$ is obtained by summing all such $\hat{x}^{(s,d)}_{m}(t)$ values corresponding to the $(s,d)$ pairs whose paths include link $\ell$, such that,



\begin{equation}
\label{link_traf}
\tau_{\ell,m}(t) = 
\sum_{(s,d)} 
\hat{x}^{(s,d)}_{m}(t),
\qquad 
\forall \, \ell \in \mathcal{P}^{(s,d)}, \; \forall m.
\end{equation}

Equation~(\ref{link_traf}) aggregates the predicted traffic contributions from all $(s,d)$ pairs whose routing paths traverse link $\ell$. Specifically, for each sequence $m$ and time step $t$, every pair $(s,d)$ contributes a fraction of the source node’s predicted traffic $\hat{x}^{(s,d)}_{m}(t)$ to the total load on link $\ell$ if $\ell \in \mathcal{P}^{(s,d)}$. Summing over all such pairs yields $\tau_{\ell,m}(t)$, which represents the cumulative predicted traffic carried by link $\ell$ at time $t$ (refer line 8 of Algorithm \ref{alg:congested_link_identification}). Such an aggregation effectively projects the node-level traffic forecasts onto the physical network links while accounting for routing topology, ensuring that links traversed by multiple $(s,d)$ paths naturally accumulate higher predicted loads, facilitating early identification of highly utilized links representing potential capacity-stressed regions of the network.
Although the uniform traffic splitting assumption may not capture the full complexity of real source-destination flows, it provides a lightweight and topology-agnostic approximation that allows us to isolate and evaluate the effectiveness of the proposed FedNET framework. Incorporating more sophisticated traffic aggregation models is left as future work.

\subsection{Link Utilization Scoring Based on Predicted Traffic}

Using the link-level traffic sequences $\tau_{\ell,m}(t)$ obtained from the node-to-link mapping step, we evaluate the \textit{link utilization score (LUS)}, a metric that quantifies the overall traffic burden on a link based on its predicted load and variability. For every link $\ell$ and sequence $m$, the mean traffic $\mu_{\ell,m}$ and standard deviation $\sigma_{\ell,m}$ are computed over the prediction horizon $p$ (refer line 12 of Algorithm \ref{alg:congested_link_identification}). The mean reflects the expected utilization of the link, while the standard deviation captures temporal variability in predicted traffic. 

To assess per-link traffic risk, we first compute the average predicted traffic statistics across all sequences:

\begin{equation}
\bar{\mu}_{\ell} = \frac{1}{|\mathcal{U}|} \sum_{m=1}^{|\mathcal{U}|} \mu_{\ell, m}
\label{eq:mu_l}
\end{equation}

\begin{equation}
\bar{\sigma}_{\ell} = \frac{1}{|\mathcal{U}|} \sum_{m=1}^{|\mathcal{U}|} \sigma_{\ell, m}
\label{eq:sigma_l}
\end{equation}

Next, both $\bar{\mu}_{\ell}$ and $\bar{\sigma}_{\ell}$ are normalized by their maximum values across all links to ensure a uniform scale for scoring. Subsequently, the LUS for each link $\ell$ is defined as:

\begin{equation}
\zeta_\ell = \beta \, \frac{\bar{\mu}_{\ell}}{\bar{\mu}_{max}} + (1-\beta) \, \frac{\bar{\sigma}_{\ell}}{\bar{\sigma}_{max}}
\label{eq:utilization_scores}
\end{equation}

\noindent where $\beta \in [0,1]$ controls the relative importance of average traffic and standard deviation. In this work, we set $\beta = 0.5$ to give equal weights to both components. The LUS $\zeta_\ell$ serves as a composite indicator of link load conditions, capturing both consistently high utilization (through $\bar{\mu}_{\ell}$) and short-term variations in traffic (through $\bar{\sigma}_{\ell}$). Links with higher $\zeta_\ell$ values are therefore more likely to experience high utilization, either due to continuous heavy usage or frequent traffic fluctuations. Alternative weighting strategies could be explored in future work to better adapt the score to specific network conditions.

All links are ranked in descending order based on their LUS $\zeta_\ell$, representing the most vulnerable links under the predicted traffic patterns. This method provides a lightweight and effective means of identifying highly utilized links in the network, even when precise source-destination flow information is unavailable, making it suitable for federated or privacy-aware environments. 

\begin{figure*}[h!]
\centering
\includegraphics[width=0.9\textwidth, height=.3\textwidth,  trim = {4cm 10cm 4cm 0cm}]{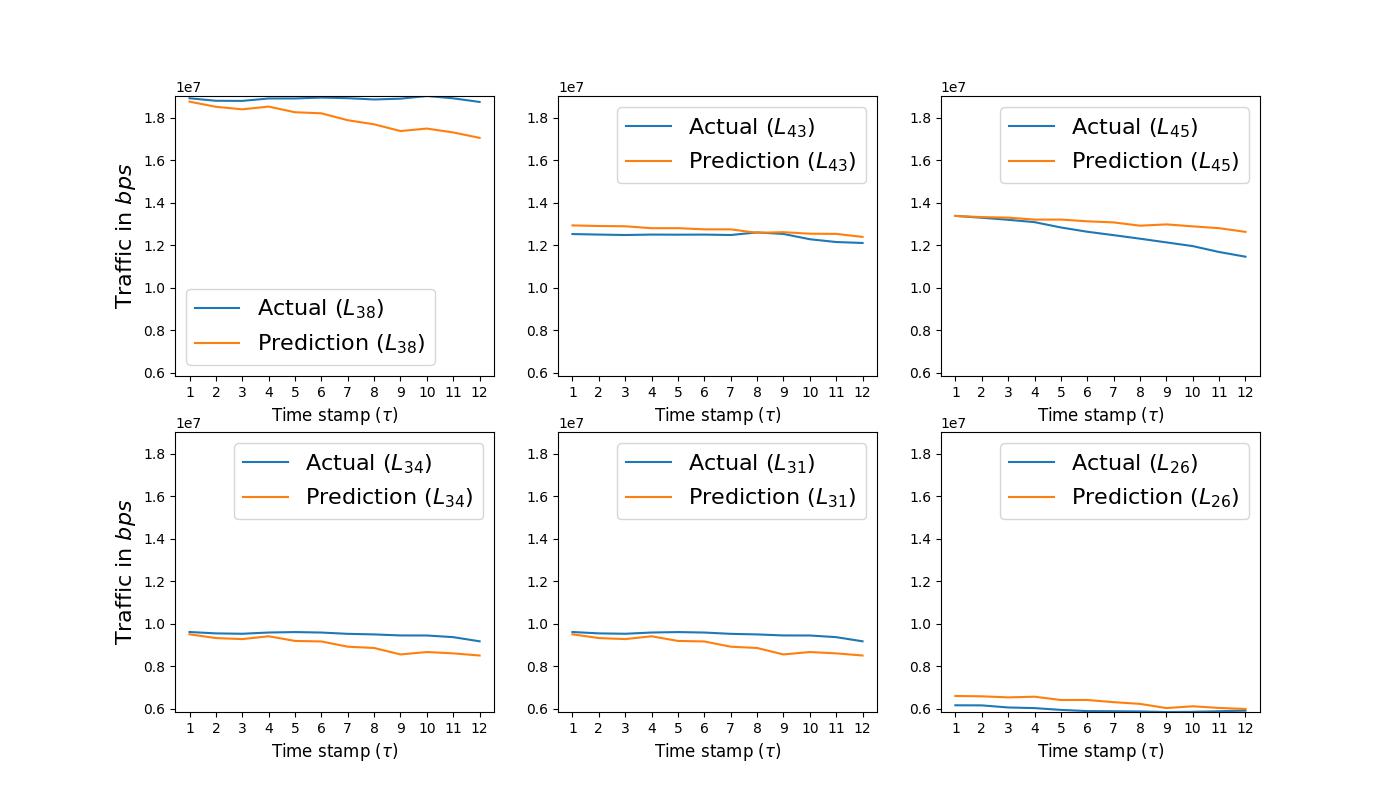}
\vspace{5cm}
\caption{The actual and the predicted traffic variation of the most utilized sequence of the top six links, selected based on the unified LUS for the longest history window and prediction horizons, i.e., $h=p=12$.}
\label{fig:congested_sequence_top_6_congested_links}
\end{figure*}

\subsection{Impact of Forecast Horizon and History Window on Link Utilization Scores}
In this section, we have analyzed two aspects of FedNET: 1) the efficacy of the proposed traffic prediction in finding the highly utilized links, 2) the effect of the  history window and prediction horizon on the LUS. So, we have undertaken a parametric analysis for different sizes of the history window and prediction horizon. We first obtained the link utilization scores of all links in the BRAIN  network topology of Fig. 1, from the unified LUS of (\ref{eq:utilization_scores}), using both the predicted throughput $\hat{\mathbf{y}}^k$ and the actual throughput $\mathbf{y}^k$. We have tabulated in Table~\ref{table:top5_links_grid} the list of the top-six most highly utilized links  (denoted $L_{\ell}$) based on the LUS obtained from the actual traffic as well as the predicted traffic. We have also presented the respective LUS. This dual presentation facilitates direct comparison and validation of the model’s predictive capability.

Across all configurations, link $L_{38}$ consistently emerges as the link with the highest LUS, identifying it as a persistent hotspot within the network. Other links such as $L_{43}$,  $L_{45}$, $L_{34}$ and  $L_{31}$ frequently appear among the top six, emphasizing their location in the network in the handling of elevated traffic loads. This consistency across different parameter settings suggests that certain links inherently exhibit higher susceptibility to future stress under varying conditions. It is worth noting that the LUS of the actual and predicted links are closely aligned, highlighting the robustness of the model’s predictions. A minor interchange is observed between the $2^{nd}$ and $3^{rd}$ most utilized links, $L_{43}$ and $L_{45}$, in the actual data compared to the predictions, which can be attributed to their nearly identical LUS. Similarly, the $4^{th}$, $5^{th}$, and $6^{th}$ positions are interchanging among $L_{31}$, $L_{34}$, $L_{68}$, and $L_{89}$. 
For the longest prediction horizon ($p=12$), a minor deviation is observed between the actual and predicted top utilized links. This arises from the accumulation of forecast errors over multiple time steps, increasing temporal uncertainty in traffic estimation.

In general, the above results show that except for some minor variations, a set of links dominates the set of the topmost utilized links, thereby demonstrating the robustness and reliability of the proposed framework even under extended forecasting horizons. It is designed to efficiently identify core bottlenecks across different forecasting conditions while remaining sensitive to variations introduced by differing history and prediction window sizes. This makes the approach well-suited for deployment in practical network environments where forecasting depth and historical context may vary across scenarios or system configurations.

Building on the above link-level utilization analysis, we now examine the dominant  patterns of the six most heavily utilized links. Each link has multiple predicted traffic sequences, so the mean and standard deviation are calculated for each sequence across time steps. Their average provides an overall measure of utilization intensity. The sequence with the highest value is then chosen as the representative utilization pattern as it effectively reflects both persistently high traffic (through the mean) and temporal fluctuations (through the standard deviation), making it a reliable indicator of potential utilization. The corresponding actual (observed) sequence is also extracted for direct comparison, allowing assessment of how well the predictions capture real high-risk behavior. Fig. ~\ref{fig:congested_sequence_top_6_congested_links} illustrates this comparison, with six subplots representing the top six links based on their unified LUS. Each subplot shows the predicted as well as the true most stress-prone sequence for the link over a prediction horizon of $\mathcal{T}=12$ time steps.

It may be observed from Fig. ~\ref{fig:congested_sequence_top_6_congested_links} that the predicted and the actual sequences in each sub-plot closely follow each other, which emphasizes the model’s effectiveness in capturing real traffic dynamics. Links $L_{34}$ and $L_{31}$ are found to carry the same set of node traffic; therefore, the node-level to link-level traffic mapping yields identical actual and predicted values for both links. It is also important to note that, in most cases, the predicted and actual values gradually diverge as we move further along the forecast horizon, indicating that the model’s performance is highly sensitive to the length of the prediction window.

Although absolute utilization scores do not represent physical traffic measurements, their relative comparison provides meaningful insights into network performance. The consistent ranking of certain links highlights potential bottleneck locations. Future work could improve this evaluation by incorporating additional operational factors to further improve high-risk link identification.

\section{Conclusions}
\label{conclusions}

In this work, we have proposed FedNET a privacy-preserving federated framework for early identification of high-risk links in large-scale networks. FedNet uses an LSTM model and FedAvg aggregation strategy for obtaining multi-step node-level traffic forecasts. It subsequently combines these forecasts with routing information to estimate the link utilization. Consequently, our method enables proactive risk assessment without requiring the sharing of sensitive data. 
Experimental results on realistic topologies demonstrate that the prediction accuracy of FedNET is comparable to that of the corresponding centralized prediction method and is inversely related to the prediction horizon. For instance, with a short prediction window ($p=1$), the framework achieves high accuracy with $R^2$ scores exceeding $0.92$ across history window sizes, while longer horizons (e.g., $p=12$) yield lower but still meaningful performance with 
$R^2$ around $0.45$–$0.55$. These findings indicate that the framework is effective for both short- and medium-term forecasting, with the best balance obtained at moderate history windows ($h=4$ or $h=8$). Thus, FedNET achieves comparable accuracy with respect to centralized methods while offering clear advantages in scalability, bandwidth efficiency, and data privacy, making it as a practical alternative for real-world deployments. 
It may, therefore, be inferred that the FL-base FedNET framework proposed in this work can be used for proactive network link utilization estimation while rendering data privacy and a low communication overhead, which makes it a potential candidate for network management in large-scale communication environments.

Future work will explore more advanced forecasting models and fairness-aware aggregation schemes to improve robustness across heterogeneous clients on larger networks. In addition, we plan to design novel link utilization scoring heuristics and integrate the framework with real-time traffic engineering for proactive network control. \\

\noindent\textbf{Disclosures:} The authors declare no conflicts of interest.
\bibliographystyle{IEEEtran}
\bibliography{IEEEabrv,TNSM_v1.bib}


\begin{IEEEbiography}
[{\includegraphics[width=1in,height=1.25in,clip,keepaspectratio]{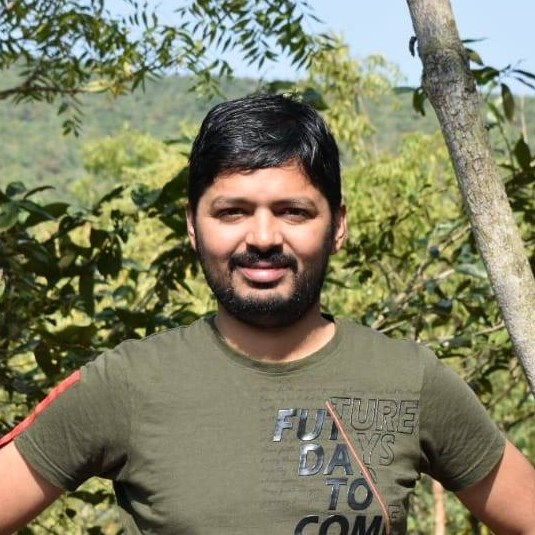}}]
{\textbf{Saroj Kumar Panda}} is currently working as a Software Professional with LTIMindtree, India. He received the M.Tech. degree from the National Institute of Technology Rourkela, India, where he is currently pursuing the Ph.D. degree in the Department of Electronics and Communication Engineering. His research interests include optical communication networks and the application of machine learning in optical networks.
\end{IEEEbiography}

\begin{IEEEbiography}
[{\includegraphics[width=1in,height=1.25in,clip,keepaspectratio]{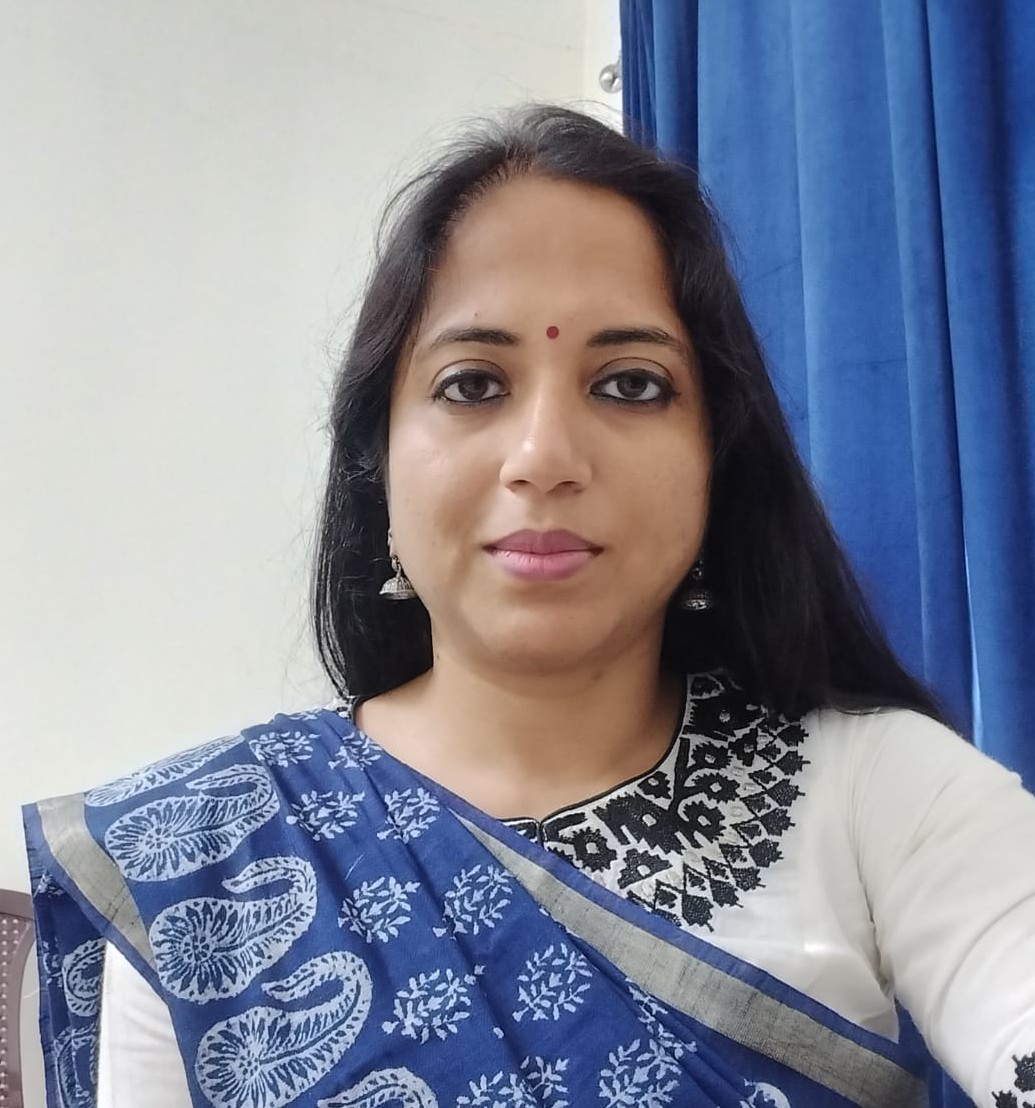}}]{Basabdatta Palit}
 has received her Ph.D. Degree from IIT Kharagpur, India in 2018. She completed her B.Tech from West Bengal University of Technology in 2009 and her M.Tech. degree in 2011 from Jadavpur University, Kolkata, India. She is currently an Assistant Professor in the Department of Electronics and Communication Engineering at National Institute of Technology, Rourkela, India. Her research interests include Cross-Layer Design and Modeling, Application of ML in Communication, Joint Sensing and Communication, URLLC, etc.
\end{IEEEbiography}

\begin{IEEEbiography}
[{\includegraphics[width=1in,height=1.25in,clip,keepaspectratio]{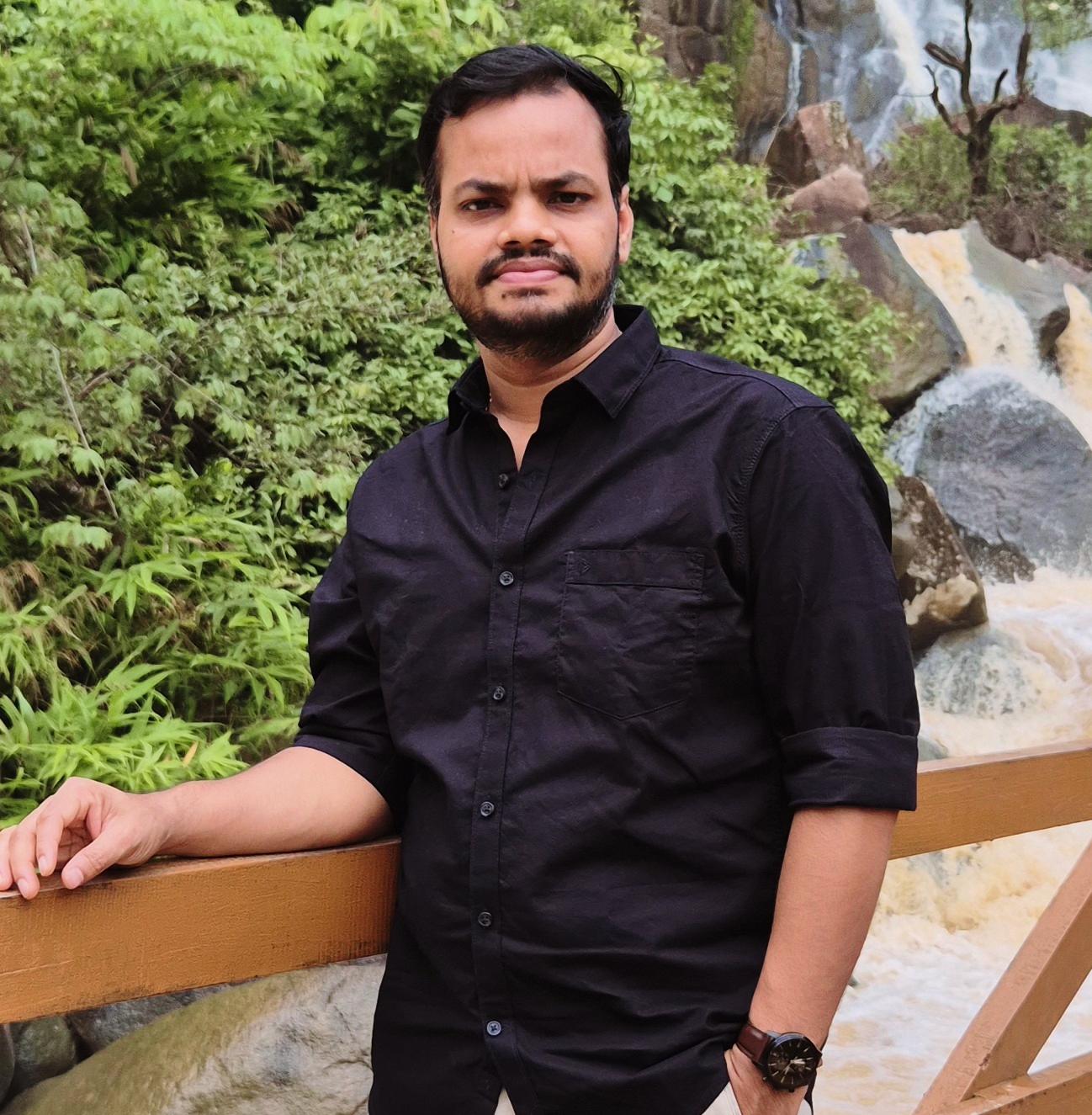}}]
{\textbf{Sadananda Behera}} received the Ph.D. degree from the Indian Institute of Technology Kharagpur, India, and the M.Tech. degree from the National Institute of Technology Rourkela, India. He is currently an Assistant Professor with the Department of Electronics and Communication Engineering, National Institute of Technology Rourkela. His research interests include optical communication networks, cross-layer optimization, resource allocation in optical networks, machine learning applications in optical networks, and quantum communication systems.
\end{IEEEbiography}

\end{document}